\documentclass[10pt,twocolumn,letterpaper]{article}

\usepackage[article]{cvpr}

\usepackage{graphicx}
\usepackage{amsmath}
\usepackage{amssymb}
\usepackage{booktabs}
\usepackage{multirow}
\usepackage{array}
\usepackage{colortbl}
\usepackage{xcolor}
\usepackage{listings}
\usepackage{enumitem}
\usepackage{algorithm}
\usepackage{algorithmic}
\usepackage{float}
\usepackage{placeins}

\definecolor{cvprblue}{rgb}{0.21,0.49,0.74}
\usepackage[pagebackref,breaklinks,colorlinks,citecolor=cvprblue,linkcolor=cvprblue,urlcolor=cvprblue]{hyperref}

\lstdefinestyle{codestyle}{
    basicstyle=\ttfamily\scriptsize,
    breaklines=true,
    frame=single,
    numbers=left,
    numberstyle=\tiny,
    tabsize=2,
    xleftmargin=1.5em,
    framexleftmargin=1.5em
}

\def\paperID{}
\def\confName{Technical Report}
\def\confYear{2026}

\begin{document}

\pagestyle{plain}
	
\title{Factorized Latent Dynamics for Video JEPA:\\An Empirical Study of Auxiliary Objectives}

\author{Santosh Premi Adhikari}
\maketitle

\begin{abstract}
Joint-Embedding Predictive Architectures (JEPA) are a promising framework for self-supervised video representation learning, yet the behavior of auxiliary objectives in small-scale Video-JEPA training is not well characterized. We report a small-scale empirical study of \textbf{18 auxiliary objective variants} for Video-JEPA across two pretraining regimes: single-dataset (UCF-101) and mixed-dataset (UCF-101 + Something-Something~V2 + ImageNet-100). We evaluate frozen representations on three complementary benchmarks: Diving-48 (fine-grained motion), Something-Something~V2 (temporal reasoning), and ImageNet-100 (appearance).

Our experiments suggest that many auxiliary objectives exhibit \textbf{capacity trade-offs}: gains on one downstream capability often coincide with degradation on another. We then study \textbf{FWM-HW-LD} (Factorized World-Model with Hard-Region-Weighted Latent Dynamics), a training-time objective that separates the latent representation into appearance and dynamics subspaces and applies hard-region weighting to both JEPA prediction errors and latent dynamics errors. In our mixed-dataset setting, FWM-HW-LD improves ImageNet-100 by \textbf{+5.92} and SSv2 by \textbf{+3.21} percentage points relative to the reference baseline, while remaining within \textbf{0.30} percentage points on Diving-48. These results indicate that latent factorization is a useful direction for studying auxiliary-objective trade-offs in Video-JEPA.
\end{abstract}

\noindent\textbf{Code:} \url{https://github.com/santoshpremi/Factorized-Latent-Dynamics-for-Video-JEPA-An-Empirical-Study-of-Auxiliary-Objectives}

\section{Introduction}
\label{sec:intro}

Video understanding requires models to capture both static appearance (what objects look like) and temporal dynamics (how they move and interact). The Video Joint-Embedding Predictive Architecture (V-JEPA)~\cite{bardes2024vjepa,assran2025vjepa2} addresses this through masked video modeling in a learned latent space, predicting embeddings of masked spatiotemporal regions from visible context without pixel-level reconstruction. This approach is theoretically motivated by the JEPA framework~\cite{lecun2022path,assran2023ijepa}, which argues that learning predictive world models in abstract latent space avoids the wasteful task of modeling unpredictable pixel-level details, in contrast to pixel-reconstruction methods such as MAE~\cite{he2022mae,tong2022videomae}.

While V-JEPA has demonstrated strong results on video understanding benchmarks~\cite{bardes2024vjepa}, the standard training objective---predicting teacher encoder embeddings with a latent prediction loss---does not explicitly isolate fine-grained temporal dynamics from appearance information. This can matter for downstream tasks such as Diving-48~\cite{li2018diving48} and Something-Something~V2~\cite{goyal2017somethingsomething}, where temporal ordering and motion cues are central.

Recent theory on JEPA with auxiliary tasks argues that auxiliary signals can determine which distinctions a representation must preserve~\cite{yu2025auxjepa}. This motivates our empirical question: under a fixed small-scale Video-JEPA budget, which auxiliary objectives help transfer, and which introduce trade-offs across motion- and appearance-sensitive evaluations?

We conduct an empirical study across \textbf{18 auxiliary objective variants}, organized into six categories:

\begin{enumerate}[topsep=2pt,itemsep=1pt,leftmargin=*]
    \item \textbf{Kinematic Regularization} --- L1, Huber, acceleration, split-channel, and annealed penalties on temporal differences.
    \item \textbf{Motion-Guided Masking} --- biasing mask sampling toward high-motion regions.
    \item \textbf{Anti-Collapse Regularization (SIGReg)} --- enforcing Gaussian-distributed latent embeddings~\cite{maes2025lejepa}.
    \item \textbf{Physics-Inspired Dynamics} --- Hamiltonian and velocity-gated priors on latent evolution.
    \item \textbf{Latent Dynamics Prediction} --- predicting temporal differences in latent space (Delta-JEPA, LD-JEPA, Spectral-JEPA, LTC-JEPA).
    \item \textbf{Factorized Latent Objectives} --- structurally separating appearance and dynamics subspaces during training (FWM-JEPA, FWM-HW-LD).
\end{enumerate}

\noindent Our contributions are as follows:

\begin{enumerate}[topsep=2pt,itemsep=1pt,leftmargin=*]
    \item A small-scale empirical study of V-JEPA auxiliary objectives, spanning 18 variants across three benchmarks under a shared training budget.
    \item Evidence for \textbf{capacity trade-offs}: many auxiliary objectives improve one downstream capability while degrading another in the same frozen encoder.
    \item \textbf{FWM-HW-LD}: a factorized latent-dynamics training objective that gives the most balanced result in our mixed-dataset sweep, improving ImageNet-100 (+5.92 pp) and SSv2 (+3.21 pp) while remaining close to the reference baseline on Diving-48 ($-0.30$ pp).
    \item Practical observations and failure cases that may guide future, larger-scale studies of auxiliary objectives for Video-JEPA.
\end{enumerate}

\begin{figure*}[t]
\centering
\includegraphics[width=\textwidth]{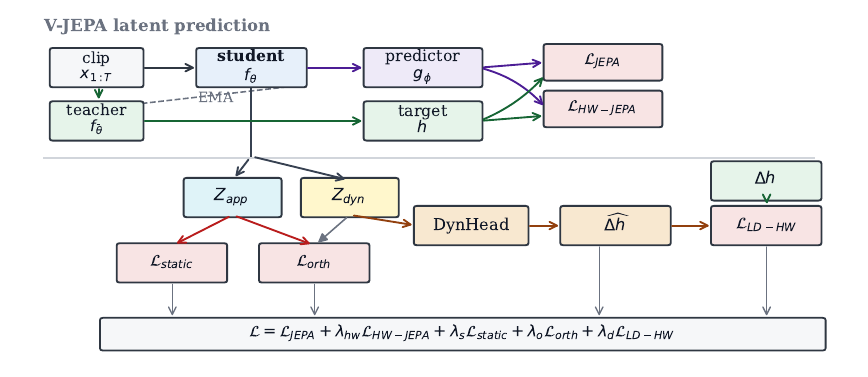}
\caption{\textbf{FWM-HW-LD training objective overview.} The standard V-JEPA latent prediction path is retained. During training, the student encoder output is additionally partitioned into appearance ($Z_{\text{app}}$) and dynamics ($Z_{\text{dyn}}$) subspaces. Hard-region weighting is applied to the JEPA prediction error ($\mathcal{L}_{\text{HW-JEPA}}$) and, when latent dynamics is enabled, to the latent dynamics error ($\mathcal{L}_{\text{LD-HW}}$). Additional FWM losses encourage temporal stability in $Z_{\text{app}}$ and decorrelation between $Z_{\text{app}}$ and $Z_{\text{dyn}}$.}
\label{fig:overview}
\end{figure*}

\section{Related Work}
\label{sec:related}

\noindent\textbf{Self-supervised image representation learning.}
Self-supervised pretext tasks for images include context prediction~\cite{doersch2015context}, jigsaw puzzles~\cite{noroozi2016jigsaw}, and contrastive instance discrimination~\cite{chen2020simclr,he2020moco,chen2020mocov2}. Cluster-based approaches such as SwAV~\cite{caron2020swav}, distillation methods such as BYOL~\cite{grill2020byol} and DINO~\cite{caron2021dino}, redundancy-reduction methods such as Barlow Twins~\cite{zbontar2021barlow}, and information-theoretic methods such as DIM~\cite{hjelm2019dim} and CPC~\cite{oord2018cpc} have all advanced the state of the art. More recently, masked image modeling methods including BEiT~\cite{bao2022beit}, SimMIM~\cite{xie2022simmim}, and MAE~\cite{he2022mae} have demonstrated that pixel reconstruction with high masking ratios learns strong visual features.

\noindent\textbf{Joint-embedding predictive architectures.}
JEPA~\cite{lecun2022path} predicts latent representations of unseen content from observed content rather than reconstructing pixels. I-JEPA~\cite{assran2023ijepa} demonstrated this paradigm at image scale, while V-JEPA~\cite{bardes2024vjepa} extended it to video. V-JEPA~2~\cite{assran2025vjepa2} scales the approach with larger backbones and curated video data, and V-JEPA~2.1~\cite{murlabadia2026vjepa21} introduces dense/context prediction, deep self-supervision, multi-modal tokenizers, and further scaling to improve dense video features while preserving global recognition. This recent work is important for our study because it shows that changing where and how the V-JEPA prediction loss is applied can improve one representation property while stressing another. Recent theory on JEPA auxiliary tasks~\cite{yu2025auxjepa} likewise suggests that auxiliary signals influence which distinctions are preserved in the latent representation. Closely related is LeJEPA~\cite{maes2025lejepa}, which proposes SIGReg, a sketched-isotropic-Gaussian regularization that prevents collapse without requiring an EMA target. Our work is orthogonal: we study which auxiliary objectives help small-scale V-JEPA learn motion-discriminative representations, on top of the underlying joint-embedding objective.

\noindent\textbf{Self-supervised video representation learning.}
Early video SSL methods exploited temporal structure as a free supervisory signal: shuffle-and-learn~\cite{misra2016shuffle}, clip order prediction~\cite{xu2019speednet}, motion segmentation from optical flow~\cite{pathak2017motion}, and cycle-contrastive prediction~\cite{kong2020cycle,han2020coclr}. Contrastive video methods such as VideoMoCo~\cite{pan2021videomoco}, BraVe~\cite{recasens2021brave}, and CBT~\cite{sun2019contrastive} extend instance discrimination to temporally augmented clips. The current state of the art is dominated by masked video modeling: VideoMAE~\cite{tong2022videomae,wang2023videomaev2}, ST-MAE~\cite{feichtenhofer2022masked}, and Siamese MAE~\cite{gupta2024siamese} reconstruct masked tubelets, while V-JEPA~\cite{bardes2024vjepa} predicts latent embeddings. Multi-view objectives~\cite{tian2020cmc,yan2022multiview} provide complementary signals.

\noindent\textbf{Video architectures and benchmarks.}
We adopt ViT~\cite{dosovitskiy2021vit,vaswani2017attention} as the encoder backbone, following V-JEPA. Alternative video backbones include I3D~\cite{carreira2017i3d}, R(2+1)D~\cite{tran2018r3d}, SlowFast~\cite{feichtenhofer2019slowfast}, MViT~\cite{fan2021mvit}, ViViT~\cite{arnab2021vivit}, TimeSformer~\cite{bertasius2021timesformer}, and Video Swin~\cite{liu2022videoswin}. Standard evaluation benchmarks include UCF-101~\cite{soomro2012ucf101} (general action recognition), Diving-48~\cite{li2018diving48} (fine-grained motion without representation bias), Something-Something~V2~\cite{goyal2017somethingsomething} (temporal reasoning), EPIC-KITCHENS~\cite{damen2022epickitchens} (egocentric), and Kinetics~\cite{carreira2017i3d}. ImageNet~\cite{russakovsky2015imagenet} provides image-level evaluation. Bardes~\etal~\cite{bardes2022vicreg} introduce variance-invariance-covariance regularization to prevent collapse.

\noindent\textbf{Latent-space factorization.}
Disentangled representation learning has been explored extensively in VAE-based frameworks~\cite{kingma2014vae,higgins2017betavae} and contrastive settings~\cite{tian2020cmc}. We apply factorization specifically to the JEPA encoder-predictor architecture, structurally separating appearance-invariant and dynamics-sensitive channel groups within the same encoder output, and combine it with hard-region weighted latent dynamics prediction. This is a training-time channel partition of V-JEPA features, not an object-centric, action-factorized, or generative world model.

\section{Method}
\label{sec:method}

\subsection{V-JEPA Baseline}

The V-JEPA~\cite{bardes2024vjepa} architecture consists of a \textbf{student encoder} $f_\theta$ (ViT-Base~\cite{dosovitskiy2021vit}, 86M parameters, patch $16\!\times\!16$, tubelet~2, RoPE positional encoding) encoding visible spatiotemporal patches from 16-frame clips into latent representations $z \in \mathbb{R}^{T \times H \times W \times D}$, a \textbf{teacher encoder} $f_{\bar{\theta}}$ (EMA of student weights, following BYOL~\cite{grill2020byol} and DINO~\cite{caron2021dino}) producing targets $h$, and a \textbf{predictor} $g_\phi$ (8-layer Transformer, embed dim~256, 8 heads) predicting masked region embeddings from visible context. Following the original V-JEPA implementation~\cite{bardes2024vjepa}, the training objective is a distance-weighted \textbf{$\ell_1$ loss}:
\begin{equation}
    \mathcal{L}_{\text{JEPA}} = \mathbb{E}\left[\|\hat{z} - \text{sg}(h)\|_1\right]
    \label{eq:jepa}
\end{equation}
where $\text{sg}(\cdot)$ denotes stop-gradient on the teacher output. Distance weighting down-scales the loss contribution of spatially distant mask regions, as in the original V-JEPA code.

\subsection{Kinematic Regularization}

Five variants penalize temporal differences in encoder embeddings. In our UCF-101 kinematic runs, EMA is disabled, so this term is applied to the trainable encoder output rather than a frozen teacher target:

\noindent\textbf{Kinematic-L1.} L1 penalty on first-order temporal differences:
\begin{equation}
    \mathcal{L}_{\text{kin}} = \lambda \cdot \mathbb{E}\left[|h^{(t+1)} - h^{(t)}|\right]
\end{equation}

\noindent\textbf{Kinematic-Huber.} Replaces L1 with Huber loss for robustness to outliers.

\noindent\textbf{Kinematic-Accel.} Penalizes second-order temporal differences:
\begin{equation}
    \mathcal{L}_{\text{accel}} = \lambda \cdot \mathbb{E}\left[|h^{(t+2)} - 2h^{(t+1)} + h^{(t)}|\right]
\end{equation}

\noindent\textbf{Kinematic-Split.} Applies the acceleration-style kinematic penalty only to the first half of the channels, leaving the remaining channels unconstrained.

\noindent\textbf{Kinematic-Anneal.} Cosine annealing of $\lambda$ throughout training.

\subsection{Motion-Guided Masking}

Standard V-JEPA masks random spatiotemporal tubes, following the masking strategies of MAE~\cite{he2022mae} and VideoMAE~\cite{tong2022videomae}. We bias mask center sampling toward high-motion regions, drawing inspiration from motion-driven SSL~\cite{pathak2017motion}:
\begin{equation}
    P(\text{center}) \propto \exp\!\left(\alpha \cdot \|\text{frame}^{(t+1)} - \text{frame}^{(t)}\|\right)
\end{equation}
With probability 0.1 we fall back to uniform random sampling. This modifies the training data distribution without adding any auxiliary loss term.

\subsection{SIGReg}

Following Maes~\etal~\cite{maes2025lejepa}, we project latent embeddings onto random unit vectors and apply normality testing. Evaluated with standard EMA teacher (\textbf{SIGReg}) and without EMA (\textbf{SIGReg-no-EMA}).

\subsection{Physics-Inspired Dynamics}

\noindent\textbf{Hamiltonian-JEPA.} Decomposes the latent embedding into position $q$ and momentum $p$ channels. A learned Hamiltonian $H(q,p)$ enforces energy-conserving dynamics via Hamilton's equations.

\noindent\textbf{Velocity-Gated JEPA (VelGate).} Applies kinematic regularization selectively to low-velocity (background) tokens only.

\subsection{Latent Dynamics Prediction}

\noindent\textbf{Delta-JEPA.} Aligns student temporal differences with teacher temporal differences:
\begin{equation}
    \mathcal{L}_{\Delta} = \|z^{(t+1)}\!-\!z^{(t)} - \text{sg}(h^{(t+1)}\!-\!h^{(t)})\|_1
\end{equation}

\noindent\textbf{LD-JEPA (Latent Dynamics).} A dedicated MLP dynamics head predicts the teacher's temporal difference from the student's representation:
\begin{equation}
    \mathcal{L}_{\text{LD}} = \|\text{DynHead}(z^{(t)}) - \text{sg}(h^{(t+1)}\!-\!h^{(t)})\|_1
\end{equation}

\noindent\textbf{Spectral-JEPA.} Applies FFT along the temporal dimension and minimizes L1 on frequency-weighted spectral coefficients.

\noindent\textbf{LTC-JEPA (Latent Temporal Contrastive).} Margin-based contrastive loss ensuring $z^{(t)}$ is closer to $h^{(t)}$ than to $h^{(t+1)}$ in cosine similarity, in the spirit of contrastive video representation learning~\cite{han2020coclr,kong2020cycle,recasens2021brave}.

\subsection{Factorized World-Model JEPA (FWM-JEPA)}

FWM-JEPA structurally separates the encoder's $D$-dimensional output into two subspaces:
\begin{itemize}[topsep=2pt,itemsep=1pt,leftmargin=*]
    \item $Z_{\text{app}}$ (first $D/2$ channels): Appearance subspace, encouraged to be temporally invariant.
    \item $Z_{\text{dyn}}$ (remaining $D/2$ channels): Dynamics subspace, free to capture temporal variation.
\end{itemize}

\begin{equation}
    \mathcal{L}_{\text{static}} = \mathbb{E}\left[|Z_{\text{app}}^{(t+1)} - Z_{\text{app}}^{(t)}|\right]
\end{equation}
\begin{equation}
    \mathcal{L}_{\text{orth}} = \frac{1}{N}\|C_{Z_{\text{app}}}^\top C_{Z_{\text{dyn}}}\|_F^2
\end{equation}
where $C$ denotes centered features.

\subsection{FWM-HW-LD: Factorized Latent-Dynamics Training Objective}

FWM-HW-LD is a training objective rather than a new inference-time architecture. It combines four loss components:

\begin{enumerate}[topsep=2pt,itemsep=1pt,leftmargin=*]
    \item \textbf{Factorization (FWM):} $Z_{\text{app}}$ and $Z_{\text{dyn}}$ are encouraged to carry less redundant information via $\mathcal{L}_{\text{static}}$ and $\mathcal{L}_{\text{orth}}$.
    \item \textbf{Hard-Weighted JEPA (HW-JEPA):} the standard JEPA prediction error is reweighted toward high-error target tokens.
    \item \textbf{Latent Dynamics (LD):} a dynamics head predicts $h^{(t+1)}\!-\!h^{(t)}$ from only $Z_{\text{dyn}}$, encouraging this subspace to carry temporally predictive information during training.
    \item \textbf{Hard-Weighted Latent Dynamics (LD-HW):} the latent dynamics error is also reweighted toward high-error temporal tokens:
\end{enumerate}

\begin{equation}
    w_i = \frac{\exp(e_i / \tau)}{\sum_j \exp(e_j / \tau)} \cdot N_{\text{tokens}}
\end{equation}
\begin{equation}
    \mathcal{L}_{\text{HW-JEPA}} = \frac{1}{N}\sum_i w_i^{\text{jepa}} \cdot \|\hat{z}_i - h_i\|_1
\end{equation}
\begin{equation}
    \mathcal{L}_{\text{LD-HW}} = \frac{1}{N}\sum_i w_i^{\text{ld}} \cdot \|\hat{\Delta}_i - \Delta h_i\|_1
\end{equation}

\noindent The total loss for FWM-HW-LD is:
\begin{equation}
    \mathcal{L} = \mathcal{L}_{\text{JEPA}} + \lambda_{hw}\mathcal{L}_{\text{HW-JEPA}} + \lambda_s \mathcal{L}_{\text{static}} + \lambda_o \mathcal{L}_{\text{orth}} + \lambda_d \mathcal{L}_{\text{LD-HW}}
    \label{eq:fwmhwld}
\end{equation}

All dynamics prediction occurs in latent space rather than pixel space, following the central design motivation of JEPA-style feature prediction. At evaluation time, we use the frozen student encoder as in the baseline; the auxiliary dynamics head is only a training-time scaffold. The objective is illustrated in Figure~\ref{fig:overview}.
The term ``factorized world-model'' in this report should be read narrowly: it refers to a split of the V-JEPA latent channels into appearance-regularized and dynamics-regularized groups during training, not to a separately learned object-factorized simulator or action-conditioned planner.

\subsection{Additional Variants Used in the Tables}

\noindent\textbf{Future-Predictive masking.} Uses the V-JEPA objective with temporally constrained complementary masks (\texttt{full\_complement=true}, \texttt{max\_temporal\_keep=0.5}) so prediction emphasizes temporally separated target regions.

\noindent\textbf{Motion-Future.} Combines Future-Predictive masking with Motion-Guided sampling (\texttt{motion\_guided=true}, strength 2.0).

\noindent\textbf{AMG-JEPA.} Aggressive Motion-Guided masking uses stronger motion bias (strength 5.0) and no random fallback (\texttt{motion\_guided\_random\_rate=0.0}).

\noindent\textbf{AC-JEPA and FAC-JEPA.} AC-JEPA predicts per-token RGB patch-mean frame differences from student features using an L1 loss. FAC-JEPA combines AC-JEPA with FWM; in that case the action head receives only $Z_{\text{dyn}}$.

\noindent\textbf{Combination rows.} AC+HW adds HW-JEPA to AC-JEPA. Combo combines Delta-JEPA with HW-JEPA and aggressive motion-guided masking. HW-LD combines HW-JEPA with hard-weighted latent dynamics but without FWM factorization.

\FloatBarrier
\section{Experimental Setup}
\label{sec:experiments}

\subsection{Pretraining Configuration}

We pretrain ViT-Base~\cite{dosovitskiy2021vit} encoders with AdamW~\cite{loshchilov2019adamw} on either UCF-101~\cite{soomro2012ucf101} alone or a balanced mixture of UCF-101, SSv2~\cite{goyal2017somethingsomething}, and ImageNet-100~\cite{russakovsky2015imagenet}. Methods share architecture, data mixture, schedule, and optimizer settings (Table~\ref{tab:config}); no per-method tuning is performed. Mixed-dataset deltas are reported as percentage-point changes against the complete three-benchmark reference baseline available for the study.

\begin{table}[H]
    \caption{Pretraining configuration for both experimental regimes.}
    \label{tab:config}
    \centering
    \scriptsize
    \setlength{\tabcolsep}{3pt}
    \begin{tabular}{lcc}
        \toprule
        \textbf{Component} & \textbf{Single-Dataset} & \textbf{Mixed-Dataset} \\
        \midrule
        Backbone & \multicolumn{2}{c}{ViT-Base (86M), patch $16\!\times\!16$, tubelet 2, RoPE} \\
        Predictor & \multicolumn{2}{c}{8-layer Transformer, embed dim 256, 8 heads} \\
        Input & \multicolumn{2}{c}{16 frames, frame stride 2, crop $224\!\times\!224$} \\
        Precision & \multicolumn{2}{c}{bfloat16} \\
        Data & UCF-101 (13K) & UCF + SSv2 + IN-100 \\
        Data weights & --- & 20\% / 60\% / 20\% \\
        Schedule & \multicolumn{2}{c}{100 epochs, 300 iter/epoch} \\
        Batch size & \multicolumn{2}{c}{$8 \times 4$ GPUs $= 32$} \\
        Optimizer & \multicolumn{2}{c}{AdamW, lr\,$=$\,$6\!\times\!10^{-4}$, wd\,$=$\,0.04} \\
        LR warmup & \multicolumn{2}{c}{1 epoch, $10^{-4} \to 6\!\times\!10^{-4}$ (cosine)} \\
        EMA momentum & \multicolumn{2}{c}{0.99925 (constant)} \\
        Loss & \multicolumn{2}{c}{$\ell_1$ (loss\_exp\,$=$\,1.0), distance-weighted} \\
        Hardware & \multicolumn{2}{c}{$4\times$ NVIDIA A100 40\,GB, $\sim$7h/run} \\
        \bottomrule
    \end{tabular}
\end{table}

\subsection{Evaluation Protocol}

We follow the frozen-encoder evaluation protocol of V-JEPA~\cite{bardes2024vjepa,assran2025vjepa2}, training attentive or linear probes on three complementary downstream benchmarks (Table~\ref{tab:eval_protocol}). Diving-48~\cite{li2018diving48} measures fine-grained motion understanding without representation bias, SSv2~\cite{goyal2017somethingsomething} measures temporal reasoning, and ImageNet-100~\cite{russakovsky2015imagenet} measures appearance recognition.

\begin{table}[H]
    \caption{Evaluation benchmarks and probe configuration.}
    \label{tab:eval_protocol}
    \centering
    \scriptsize
    \setlength{\tabcolsep}{3pt}
    \begin{tabular}{llccl}
        \toprule
        \textbf{Benchmark} & \textbf{Task} & \textbf{Cls.} & \textbf{Ep.} & \textbf{Probe} \\
        \midrule
        Diving-48 & Fine-grained diving & 48 & 50 & Attentive \\
        ImageNet-100 & Object/scene recog. & 100 & 50 & Linear \\
        SSv2 & Temporal reasoning & 174 & 50 & Attentive \\
        \bottomrule
    \end{tabular}
\end{table}

\section{Results}
\label{sec:results}

\subsection{Single-Dataset Results (UCF-101 Pretraining)}

Table~\ref{tab:ucf_results} presents results from 12 methods pretrained on UCF-101 only.

\begin{table}[t]
    \caption{Top-1 accuracy (\%) with UCF-101 pretraining. $\uparrow$/$\downarrow$ denote percentage-point changes relative to baseline. \textbf{Bold} = best. Motion-Guided improves all reported metrics in this table.}
    \label{tab:ucf_results}
    \centering
    \scriptsize
    \setlength{\tabcolsep}{3pt}
    \begin{tabular}{lcccl}
        \toprule
        \textbf{Method} & \textbf{D-48} & \textbf{IN-100} & \textbf{SSv2} & \textbf{Univ.} \\
        \midrule
        Baseline V-JEPA & 8.38 & 12.02 & 2.07 & --- \\
        \textbf{Motion-Guided} & \textbf{8.68}{\scriptsize$\uparrow$} & \textbf{12.16}{\scriptsize$\uparrow$} & \textbf{3.45}{\scriptsize$\uparrow$} & \textbf{Yes} \\
        \midrule
        Kin.-L1 ($\lambda$=1.0) & 5.58{\scriptsize$\downarrow$} & 13.64{\scriptsize$\uparrow$} & --- & No \\
        Kin.-L1 ($\lambda$=0.1) & 5.84{\scriptsize$\downarrow$} & 13.66{\scriptsize$\uparrow$} & 3.07{\scriptsize$\uparrow$} & No \\
        Kin.-Accel & 5.79{\scriptsize$\downarrow$} & 13.74{\scriptsize$\uparrow$} & 3.18{\scriptsize$\uparrow$} & No \\
        Kin.-Huber & 5.99{\scriptsize$\downarrow$} & --- & --- & No \\
        Kin.-Split & 5.48{\scriptsize$\downarrow$} & 13.58{\scriptsize$\uparrow$} & --- & No \\
        Kin.-Anneal & 5.63{\scriptsize$\downarrow$} & --- & --- & No \\
        \midrule
        SIGReg & 6.29{\scriptsize$\downarrow$} & --- & --- & No \\
        SIGReg-no-EMA & 5.99{\scriptsize$\downarrow$} & --- & --- & No \\
        Hamiltonian & 6.04{\scriptsize$\downarrow$} & --- & --- & No \\
        VelGate & 5.84{\scriptsize$\downarrow$} & --- & --- & No \\
        Future-Predictive & 7.11{\scriptsize$\downarrow$} & --- & --- & No \\
        Motion-Future & 6.45{\scriptsize$\downarrow$} & --- & --- & No \\
        \bottomrule
    \end{tabular}
\end{table}

\noindent\textbf{Finding~1:} Motion-Guided Masking improves all reported metrics in the UCF-101 setting (+0.30 pp D-48, +0.14 pp IN-100, +1.38 pp SSv2). Kinematic variants degrade D-48 by 2.5--2.9 points while improving IN-100 by 1.5--1.7 points, suggesting a motion--appearance trade-off under this training budget.

\subsection{Mixed-Dataset Results}

Table~\ref{tab:mixed_results} presents the complete mixed-dataset results.

\begin{table*}[t]
    \caption{Top-1 accuracy (\%) with mixed-dataset pretraining (UCF-101 + SSv2 + ImageNet-100). $\Delta$ is the percentage-point change relative to the complete reference baseline used in this study. \colorbox{green!15}{\textbf{Green}} = gains. \colorbox{red!10}{Red} = losses $>$5 points.}
    \label{tab:mixed_results}
    \centering
    \small
    \setlength{\tabcolsep}{5pt}
    \begin{tabular}{lcccccc}
        \toprule
        \textbf{Method} & \textbf{Diving-48} & $\Delta$ & \textbf{ImageNet-100} & $\Delta$ & \textbf{SSv2} & $\Delta$ \\
        \midrule
        Baseline & 8.68 & --- & 24.86 & --- & 8.39 & --- \\
        \midrule
        \rowcolor{green!8}
        \textbf{FWM-HW-LD} & \textbf{8.38} & $-$0.30 & \textbf{30.78} & \cellcolor{green!15}\textbf{+5.92} & \textbf{11.60} & \cellcolor{green!15}\textbf{+3.21} \\
        LD-JEPA & 6.95 & $-$1.73 & 22.80 & $-$2.06 & 13.41 & \cellcolor{green!15}+5.02 \\
        FWM-JEPA & 7.16 & $-$1.52 & 26.74 & \cellcolor{green!15}+1.88 & 7.00 & $-$1.39 \\
        AC+HW-JEPA & 9.39 & \cellcolor{green!15}+0.71 & 18.14 & \cellcolor{red!10}$-$6.72 & 4.88 & $-$3.51 \\
        AMG-JEPA & 5.89 & $-$2.79 & 23.54 & $-$1.32 & 7.78 & $-$0.61 \\
        HW-JEPA & 8.98 & +0.30 & 18.46 & \cellcolor{red!10}$-$6.40 & 6.30 & $-$2.09 \\
        HW-LD-JEPA & 7.92 & $-$0.76 & 19.78 & \cellcolor{red!10}$-$5.08 & 5.41 & $-$2.98 \\
        FWM-LD-JEPA & 6.40 & $-$2.28 & 14.72 & \cellcolor{red!10}$-$10.14 & 7.63 & $-$0.76 \\
        Combo & 7.16 & $-$1.52 & 18.60 & \cellcolor{red!10}$-$6.26 & 5.00 & $-$3.39 \\
        Delta-JEPA & 7.41 & $-$1.27 & 16.76 & \cellcolor{red!10}$-$8.10 & 3.80 & $-$4.59 \\
        AC-JEPA & 8.38 & $-$0.30 & 11.70 & \cellcolor{red!10}$-$13.16 & 4.21 & $-$4.18 \\
        FAC-JEPA & 8.38 & $-$0.30 & 8.58 & \cellcolor{red!10}$-$16.28 & 3.08 & \cellcolor{red!10}$-$5.31 \\
        LTC-JEPA & 7.77 & $-$0.91 & 11.34 & \cellcolor{red!10}$-$13.52 & 4.32 & $-$4.07 \\
        Spectral-JEPA & 5.69 & $-$2.99 & 12.62 & \cellcolor{red!10}$-$12.24 & 3.37 & \cellcolor{red!10}$-$5.02 \\
        \bottomrule
    \end{tabular}
\end{table*}

\noindent\textbf{Finding~2:} FWM-HW-LD achieves +5.92 percentage points on ImageNet-100 and +3.21 percentage points on SSv2 while remaining close to the Diving-48 reference baseline ($-$0.30 percentage points). In this mixed-dataset sweep, it gives the most balanced result among the tested variants, but this should be interpreted as a single-seed empirical signal rather than a statistically established ranking.

\noindent\textit{Implementation note.} The HW coefficient affects two terms when LD is enabled: it adds $\mathcal{L}_{\text{HW-JEPA}}$ on the standard prediction error, and it also hard-weights the latent dynamics error in $\mathcal{L}_{\text{LD-HW}}$.

\noindent\textbf{Finding~3:} LD-JEPA achieves +5.02 pp on SSv2, the largest temporal reasoning gain in the table, suggesting that latent dynamics prediction can help temporal reasoning in this setup even when it hurts other benchmarks.

\noindent\textbf{Finding~4:} Many auxiliary objectives coincide with substantial degradation---10 of 14 methods lose $>$5 points on ImageNet-100, and pixel-prediction objectives (AC-JEPA, FAC-JEPA) are particularly weak in this setup ($-$13 to $-$16 pp).

\subsection{Ablation: Components of FWM-HW-LD}

\begin{table}[t]
    \caption{Ablation of FWM-HW-LD components on mixed-dataset pretraining. ``HW+LD'' and ``FWM+HW+LD'' include both hard-weighted JEPA prediction and hard-weighted latent dynamics.}
    \label{tab:ablation}
    \centering
    \scriptsize
    \setlength{\tabcolsep}{4pt}
    \begin{tabular}{lccc}
        \toprule
        \textbf{Components} & \textbf{D-48} & \textbf{IN-100} & \textbf{SSv2} \\
        \midrule
        Baseline & 8.68 & 24.86 & 8.39 \\
        \midrule
        LD only & 6.95 & 22.80 & \textbf{13.41} \\
        HW + LD & 7.92 & 19.78 & 5.41 \\
        FWM only & 7.16 & 26.74 & 7.00 \\
        FWM + LD & 6.40 & 14.72 & 7.63 \\
        \rowcolor{green!8}
        \textbf{FWM + HW + LD} & \textbf{8.38} & \textbf{30.78} & 11.60 \\
        \bottomrule
    \end{tabular}
\end{table}

The ablation (Table~\ref{tab:ablation}) suggests that the components interact non-additively. LD alone boosts SSv2 ($+5.02$) but hurts ImageNet and Diving-48. FWM alone boosts ImageNet ($+1.88$) but hurts SSv2 and Diving-48. FWM+LD without hard weighting performs poorly on ImageNet ($-10.14$). The full FWM-HW-LD combination gives the most balanced result in this ablation, consistent with the hypothesis that FWM, LD, and HW are complementary under this training budget.

\subsection{Synthetic Motion Discrimination}

\begin{table}[t]
    \caption{8-class synthetic motion task (translation, rotation, scaling).}
    \label{tab:synthetic}
    \centering
    \scriptsize
    \begin{tabular}{lc}
        \toprule
        \textbf{Method} & \textbf{Accuracy (\%)} \\
        \midrule
        Baseline & 27.5 \\
        Kinematic-L1 ($\lambda$=1.0) & 67.5 \\
        Kinematic-L1 ($\lambda$=0.1) & 72.5 \\
        \bottomrule
    \end{tabular}
\end{table}

The +40--45 point improvement (Table~\ref{tab:synthetic}) confirms kinematic regularization encodes strong temporal structure, but this structure suits synthetic primitives rather than real-world fine-grained motion.

\section{Analysis}
\label{sec:analysis}

\begin{figure*}[t]
\centering
\includegraphics[width=\textwidth]{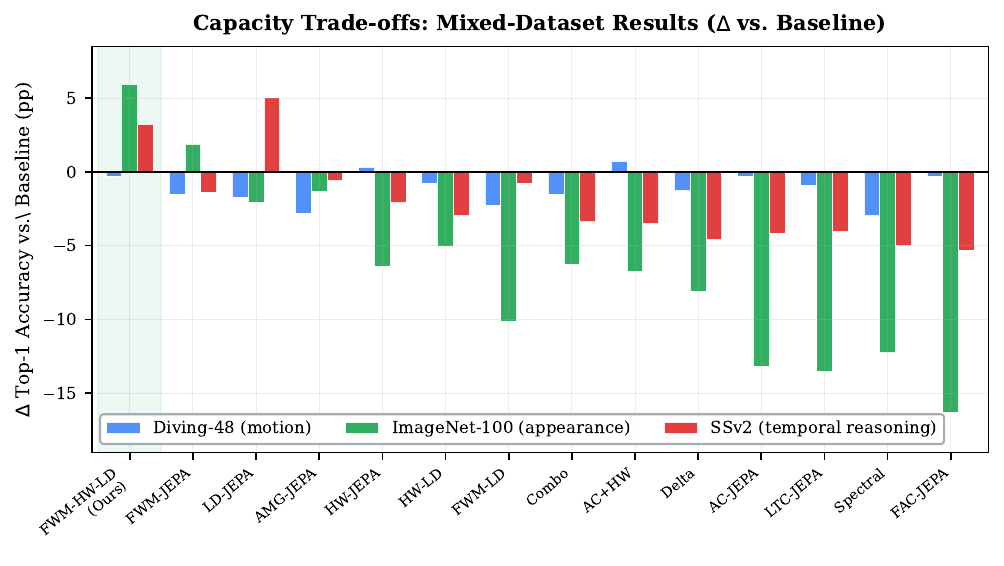}
\caption{\textbf{Capacity trade-offs across mixed-dataset methods.} Grouped bars show percentage-point $\Delta$ accuracy relative to baseline on three benchmarks. FWM-HW-LD (leftmost, highlighted) improves appearance (ImageNet-100, green) and temporal reasoning (SSv2, red) while staying close to the Diving-48 baseline. Pixel-prediction methods (AC-JEPA, FAC-JEPA) perform poorly on ImageNet-100 in this setting.}
\label{fig:interference}
\end{figure*}

\begin{figure*}[t]
\centering
\includegraphics[width=\textwidth]{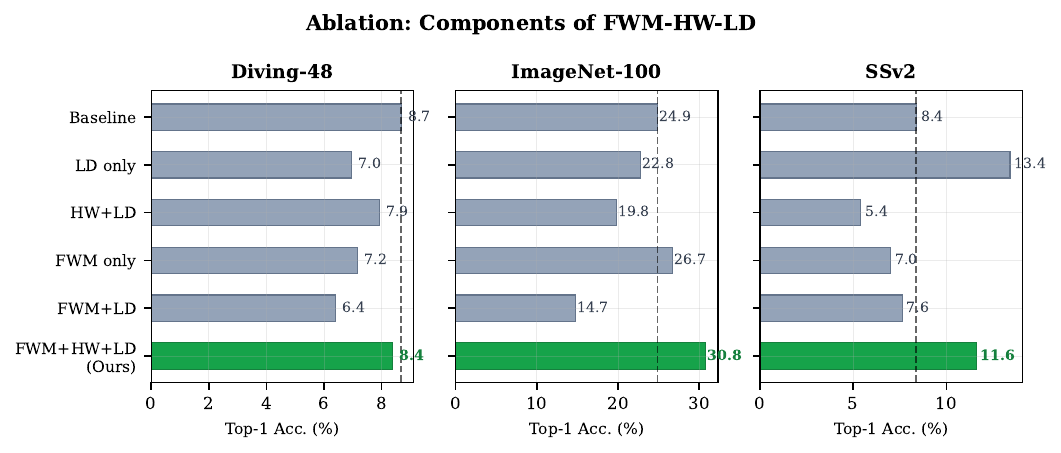}
\caption{\textbf{Ablation of FWM-HW-LD components.} Dashed lines indicate the baseline. The tested components interact non-additively: LD alone boosts SSv2 but hurts others; FWM alone boosts ImageNet-100 modestly; FWM+LD without hard weighting performs poorly on ImageNet-100. The full FWM+HW+LD combination (green bars) gives the most balanced result in this single-seed ablation.}
\label{fig:ablation}
\end{figure*}

\subsection{Capacity Trade-Offs in a Shared Latent Space}

The encoder produces a fixed 768-dimensional embedding that must simultaneously encode (1) what objects are present and (2) how they move. In our experiments, auxiliary objectives that emphasize temporal structure often coincide with weaker appearance discrimination. This is evidenced by trade-off patterns: kinematic variants gain +1.5--1.7 pp on ImageNet but lose $-$2.5--2.9 pp on Diving-48; action-conditioning variants preserve Diving-48 but substantially degrade appearance.

\subsection{Why Factorization May Help}

FWM-HW-LD may help by structurally partitioning the latent space during training. $Z_{\text{app}}$ is encouraged to remain temporally stable via $\mathcal{L}_{\text{static}}$, while $Z_{\text{dyn}}$ receives the latent dynamics prediction signal. The orthogonality constraint discourages the two subspaces from encoding redundant information. This interpretation is consistent with the observed ablation trends, but it should be validated with multi-seed runs and representation analysis.

\subsection{Pixel Prediction Is Weak in This Setup}

AC-JEPA ($-$13.16 pp IN-100) and FAC-JEPA ($-$16.28 pp) perform poorly on ImageNet-100. A plausible explanation is that predicting pixel-level RGB deltas encourages sensitivity to low-level changes that are not useful for frozen semantic recognition. This is consistent with the JEPA motivation for feature-space prediction, though it does not prove that all pixel-space auxiliaries are harmful.

\subsection{Hard-Region Weighting as an Empirical Stabilizer}

In this ablation, hard-region weighting appears important: without it (FWM+LD), ImageNet drops to 14.72\%, while the full method reaches 30.78\%. We hypothesize that hard-region weighting changes the balance of gradients by focusing the dynamics loss on tokens where temporal prediction error is highest. This is an empirical observation from this sweep, not yet a mechanistic proof.

\section{Discussion}
\label{sec:discussion}

\subsection{Practical Observations}

\begin{enumerate}[topsep=2pt,itemsep=1pt,leftmargin=*]
    \item \textbf{Mixed-data encoders:} FWM-HW-LD is the most balanced variant in our mixed-dataset runs, especially when ImageNet-100 and SSv2 are both important.
    \item \textbf{Single-dataset pretraining:} Motion-Guided Masking is a simple low-risk baseline worth testing, since it improves all reported UCF-pretraining metrics in our runs.
    \item \textbf{Temporal reasoning priority:} LD-JEPA gives the largest SSv2 gain, but with clear degradation on ImageNet-100 and Diving-48.
    \item \textbf{Caution:} Pixel-space prediction objectives (AC-JEPA, FAC-JEPA) and several frequency/contrastive objectives hurt transfer in this implementation.
\end{enumerate}

\subsection{Limitations and Threats to Validity}

This report should be read as a small-scale empirical study. Mixed-dataset experiments are single-seed, so statistical significance remains unknown. The complete three-benchmark mixed baseline used for deltas is a reference run rather than a same-seed replicate for every auxiliary method; the deltas should therefore be interpreted as empirical signals, not precise confidence intervals. FWM-HW-LD also increases training-time complexity by adding auxiliary losses and a dynamics head, although downstream evaluation still uses the frozen student encoder. Pretraining uses UCF-101 + SSv2 + ImageNet-100 rather than the large-scale data used by V-JEPA~2.x; conclusions concern relative objective behavior under this budget, not parity with large-scale models. Three evaluation benchmarks cannot cover all downstream tasks. Hyperparameters ($\lambda_s$, $\lambda_o$, $\lambda_d$, $\lambda_{hw}$) were selected conservatively but not exhaustively tuned per method. Finally, although recent JEPA theory supports the view that auxiliary signals shape which distinctions representations preserve~\cite{yu2025auxjepa}, our capacity-trade-off interpretation is based on downstream accuracy patterns; stronger evidence would require multi-seed runs, representation diagnostics, and larger-scale replication.

\section{Conclusion}
\label{sec:conclusion}

We presented a small-scale empirical study of auxiliary objectives for Video-JEPA, evaluating 18 variants across three complementary benchmarks under two pretraining regimes. Our results suggest that auxiliary objectives often expose a trade-off between appearance and dynamics-sensitive behavior in a shared latent representation.

We studied FWM-HW-LD, a training-time objective that structurally separates the latent space and applies focused latent-space dynamics prediction to hard regions. In our mixed-dataset setting, FWM-HW-LD improves ImageNet-100 by +5.92 and SSv2 by +3.21 percentage points, while remaining within 0.30 percentage points of the Diving-48 baseline.

These results suggest that latent-space organization is a useful direction for future Video-JEPA work, but the present evidence is preliminary. Rather than claiming a solved objective, we view factorized latent dynamics as a controlled way to study how appearance and motion-sensitive information compete or cooperate inside a frozen video representation.

\vspace{4pt}
\noindent\textbf{Acknowledgments.} I thank the Computer Vision Lab, CAIDAS \& IFI, University of W\"urzburg, Germany, for the research environment and computing support.

{\small
\bibliographystyle{ieeenat_fullname}
\bibliography{references}

\begin{thebibliography}{53}
\providecommand{\natexlab}[1]{#1}
\providecommand{\url}[1]{\texttt{#1}}
\expandafter\ifx\csname urlstyle\endcsname\relax
  \providecommand{\doi}[1]{doi: #1}\else
  \providecommand{\doi}{doi: \begingroup \urlstyle{rm}\Url}\fi

\bibitem[Arnab et~al.(2021)Arnab, Dehghani, Heigold, Sun, Lu{\v{c}}i{\'c}, and
  Schmid]{arnab2021vivit}
Anurag Arnab, Mostafa Dehghani, Georg Heigold, Chen Sun, Mario Lu{\v{c}}i{\'c},
  and Cordelia Schmid.
\newblock {ViViT}: A video vision transformer.
\newblock In \emph{ICCV}, 2021.

\bibitem[Assran et~al.(2023)Assran, Duval, Misra, Bojanowski, Vincent, Rabbat,
  LeCun, and Ballas]{assran2023ijepa}
Mahmoud Assran, Quentin Duval, Ishan Misra, Piotr Bojanowski, Pascal Vincent,
  Michael Rabbat, Yann LeCun, and Nicolas Ballas.
\newblock Self-supervised learning from images with a joint-embedding
  predictive architecture.
\newblock In \emph{CVPR}, 2023.

\bibitem[Assran et~al.(2025)Assran, Bardes, Fan, Garrido, Howes, Komeili,
  Muckley, Rizvi, Roberts, Sinha, Zholus, LeCun, Rabbat, and
  Ballas]{assran2025vjepa2}
Mahmoud Assran, Adrien Bardes, David Fan, Quentin Garrido, Russell Howes,
  Mojtaba Komeili, Matthew Muckley, Ammar Rizvi, Claire Roberts, Koustuv Sinha,
  Artem Zholus, Yann LeCun, Michael Rabbat, and Nicolas Ballas.
\newblock {V-JEPA 2}: Self-supervised video models enable understanding,
  prediction and planning.
\newblock \emph{arXiv preprint arXiv:2506.09985}, 2025.

\bibitem[Bao et~al.(2022)Bao, Dong, Piao, and Wei]{bao2022beit}
Hangbo Bao, Li Dong, Songhao Piao, and Furu Wei.
\newblock {BEiT}: Bert pre-training of image transformers.
\newblock In \emph{ICLR}, 2022.

\bibitem[Bardes et~al.(2022)Bardes, Ponce, and LeCun]{bardes2022vicreg}
Adrien Bardes, Jean Ponce, and Yann LeCun.
\newblock {VICReg}: Variance-invariance-covariance regularization for
  self-supervised learning.
\newblock In \emph{ICLR}, 2022.

\bibitem[Bardes et~al.(2024)Bardes, Garrido, Ponce, Chen, Rabbat, LeCun,
  Assran, and Ballas]{bardes2024vjepa}
Adrien Bardes, Quentin Garrido, Jean Ponce, Xinlei Chen, Michael Rabbat, Yann
  LeCun, Mahmoud Assran, and Nicolas Ballas.
\newblock {V-JEPA}: Video joint embedding predictive architecture.
\newblock \emph{arXiv preprint arXiv:2404.08471}, 2024.

\bibitem[Bertasius et~al.(2021)Bertasius, Wang, and
  Torresani]{bertasius2021timesformer}
Gedas Bertasius, Heng Wang, and Lorenzo Torresani.
\newblock Is space-time attention all you need for video understanding?
\newblock In \emph{ICML}, 2021.

\bibitem[Caron et~al.(2020)Caron, Misra, Mairal, Goyal, Bojanowski, and
  Joulin]{caron2020swav}
Mathilde Caron, Ishan Misra, Julien Mairal, Priya Goyal, Piotr Bojanowski, and
  Armand Joulin.
\newblock Unsupervised learning of visual features by contrasting cluster
  assignments.
\newblock In \emph{NeurIPS}, 2020.

\bibitem[Caron et~al.(2021)Caron, Touvron, Misra, J{\'e}gou, Mairal,
  Bojanowski, and Joulin]{caron2021dino}
Mathilde Caron, Hugo Touvron, Ishan Misra, Herv{\'e} J{\'e}gou, Julien Mairal,
  Piotr Bojanowski, and Armand Joulin.
\newblock Emerging properties in self-supervised vision transformers.
\newblock In \emph{ICCV}, 2021.

\bibitem[Carreira and Zisserman(2017)]{carreira2017i3d}
Jo{\~a}o Carreira and Andrew Zisserman.
\newblock Quo vadis, action recognition? a new model and the kinetics dataset.
\newblock In \emph{CVPR}, 2017.

\bibitem[Chen et~al.(2020{\natexlab{a}})Chen, Kornblith, Norouzi, and
  Hinton]{chen2020simclr}
Ting Chen, Simon Kornblith, Mohammad Norouzi, and Geoffrey Hinton.
\newblock A simple framework for contrastive learning of visual
  representations.
\newblock In \emph{ICML}, 2020{\natexlab{a}}.

\bibitem[Chen et~al.(2020{\natexlab{b}})Chen, Fan, Girshick, and
  He]{chen2020mocov2}
Xinlei Chen, Haoqi Fan, Ross Girshick, and Kaiming He.
\newblock Improved baselines with momentum contrastive learning,
  2020{\natexlab{b}}.

\bibitem[Damen et~al.(2022)Damen, Doughty, Farinella, Furnari, Ma, Kazakos,
  Moltisanti, Munro, Perrett, Price, and Wray]{damen2022epickitchens}
Dima Damen, Hazel Doughty, Giovanni~Maria Farinella, Antonino Furnari, Jian Ma,
  Evangelos Kazakos, Davide Moltisanti, Jonathan Munro, Toby Perrett, Will
  Price, and Michael Wray.
\newblock Rescaling egocentric vision: Collection, pipeline and challenges for
  {EPIC-KITCHENS-100}.
\newblock \emph{International Journal of Computer Vision}, 2022.

\bibitem[Doersch et~al.(2015)Doersch, Gupta, and Efros]{doersch2015context}
Carl Doersch, Abhinav Gupta, and Alexei~A. Efros.
\newblock Unsupervised visual representation learning by context prediction.
\newblock In \emph{ICCV}, 2015.

\bibitem[Dosovitskiy et~al.(2021)Dosovitskiy, Beyer, Kolesnikov, Weissenborn,
  Zhai, Unterthiner, Dehghani, Minderer, Heigold, Gelly, Uszkoreit, and
  Houlsby]{dosovitskiy2021vit}
Alexey Dosovitskiy, Lucas Beyer, Alexander Kolesnikov, Dirk Weissenborn,
  Xiaohua Zhai, Thomas Unterthiner, Mostafa Dehghani, Matthias Minderer, Georg
  Heigold, Sylvain Gelly, Jakob Uszkoreit, and Neil Houlsby.
\newblock An image is worth 16x16 words: Transformers for image recognition at
  scale.
\newblock In \emph{ICLR}, 2021.

\bibitem[Fan et~al.(2021)Fan, Xiong, Mangalam, Li, Yan, Malik, and
  Feichtenhofer]{fan2021mvit}
Haoqi Fan, Bo Xiong, Karttikeya Mangalam, Yanghao Li, Zhicheng Yan, Jitendra
  Malik, and Christoph Feichtenhofer.
\newblock Multiscale vision transformers.
\newblock In \emph{ICCV}, 2021.

\bibitem[Feichtenhofer et~al.(2019)Feichtenhofer, Fan, Malik, and
  He]{feichtenhofer2019slowfast}
Christoph Feichtenhofer, Haoqi Fan, Jitendra Malik, and Kaiming He.
\newblock Slowfast networks for video recognition.
\newblock In \emph{ICCV}, 2019.

\bibitem[Feichtenhofer et~al.(2022)Feichtenhofer, Fan, Li, and
  He]{feichtenhofer2022masked}
Christoph Feichtenhofer, Haoqi Fan, Yanghao Li, and Kaiming He.
\newblock Masked autoencoders as spatiotemporal learners.
\newblock In \emph{NeurIPS}, 2022.

\bibitem[Goyal et~al.(2017)Goyal, Ebrahimi~Kahou, Michalski, Materzynska,
  Westphal, Kim, Haenel, Fruend, Yianilos, Mueller-Freitag, Hoppe, Thurau, Bax,
  and Memisevic]{goyal2017somethingsomething}
Raghav Goyal, Samira Ebrahimi~Kahou, Vincent Michalski, Joanna Materzynska,
  Susanne Westphal, Heuna Kim, Valentin Haenel, Ingo Fruend, Peter Yianilos,
  Moritz Mueller-Freitag, Florian Hoppe, Christian Thurau, Ingo Bax, and Roland
  Memisevic.
\newblock The ``something something'' video database for learning and
  evaluating visual common sense.
\newblock In \emph{ICCV}, 2017.

\bibitem[Grill et~al.(2020)Grill, Strub, Altch{\'e}, Tallec, Richemond,
  Buchatskaya, Doersch, Pires, Guo, Azar, Piot, Kavukcuoglu, Munos, and
  Valko]{grill2020byol}
Jean-Bastien Grill, Florian Strub, Florent Altch{\'e}, Corentin Tallec,
  Pierre~H. Richemond, Elena Buchatskaya, Carl Doersch, Bernardo~Avila Pires,
  Zhaohan~Daniel Guo, Mohammad~Gheshlaghi Azar, Bilal Piot, Koray Kavukcuoglu,
  R{\'e}mi Munos, and Michal Valko.
\newblock Bootstrap your own latent: A new approach to self-supervised
  learning.
\newblock In \emph{NeurIPS}, 2020.

\bibitem[Gupta et~al.(2024)Gupta, Yu, Sohn, Gu, Hahn, Li, Essa, Jiang, and
  Lezama]{gupta2024siamese}
Agrim Gupta, Lijun Yu, Kihyuk Sohn, Xiuye Gu, Meera Hahn, Fei-Fei Li, Irfan
  Essa, Lu Jiang, and Jos{\'e} Lezama.
\newblock Siamese masked autoencoders.
\newblock In \emph{NeurIPS}, 2024.

\bibitem[Han et~al.(2020)Han, Xie, and Zisserman]{han2020coclr}
Tengda Han, Weidi Xie, and Andrew Zisserman.
\newblock Self-supervised co-training for video representation learning.
\newblock In \emph{NeurIPS}, 2020.

\bibitem[He et~al.(2020)He, Fan, Wu, Xie, and Girshick]{he2020moco}
Kaiming He, Haoqi Fan, Yuxin Wu, Saining Xie, and Ross Girshick.
\newblock Momentum contrast for unsupervised visual representation learning.
\newblock In \emph{CVPR}, 2020.

\bibitem[He et~al.(2022)He, Chen, Xie, Li, Doll{\'a}r, and Girshick]{he2022mae}
Kaiming He, Xinlei Chen, Saining Xie, Yanghao Li, Piotr Doll{\'a}r, and Ross
  Girshick.
\newblock Masked autoencoders are scalable vision learners.
\newblock In \emph{CVPR}, 2022.

\bibitem[Higgins et~al.(2017)Higgins, Matthey, Pal, Burgess, Glorot, Botvinick,
  Mohamed, and Lerchner]{higgins2017betavae}
Irina Higgins, Loic Matthey, Arka Pal, Christopher Burgess, Xavier Glorot,
  Matthew Botvinick, Shakir Mohamed, and Alexander Lerchner.
\newblock $\beta$-{VAE}: Learning basic visual concepts with a constrained
  variational framework.
\newblock In \emph{ICLR}, 2017.

\bibitem[Hjelm et~al.(2019)Hjelm, Fedorov, Lavoie-Marchildon, Grewal, Bachman,
  Trischler, and Bengio]{hjelm2019dim}
R~Devon Hjelm, Alex Fedorov, Samuel Lavoie-Marchildon, Karan Grewal, Phil
  Bachman, Adam Trischler, and Yoshua Bengio.
\newblock Learning deep representations by mutual information estimation and
  maximization.
\newblock In \emph{ICLR}, 2019.

\bibitem[Kingma and Welling(2014)]{kingma2014vae}
Diederik~P. Kingma and Max Welling.
\newblock Auto-encoding variational bayes.
\newblock In \emph{ICLR}, 2014.

\bibitem[Kong et~al.(2020)Kong, Wei, Deng, Yoshinaga, and
  Murakami]{kong2020cycle}
Quan Kong, Wen Wei, Ziwei Deng, Tomoo Yoshinaga, and Tomokazu Murakami.
\newblock Cycle-contrast for self-supervised video representation learning.
\newblock In \emph{NeurIPS}, 2020.

\bibitem[LeCun(2022)]{lecun2022path}
Yann LeCun.
\newblock A path towards autonomous machine intelligence, 2022.
\newblock Open review essay.

\bibitem[Li et~al.(2018)Li, Li, and Vasconcelos]{li2018diving48}
Yingwei Li, Yi Li, and Nuno Vasconcelos.
\newblock Resound: Towards action recognition without representation bias.
\newblock In \emph{ECCV}, 2018.

\bibitem[Liu et~al.(2022)Liu, Ning, Cao, Wei, Zhang, Lin, and
  Hu]{liu2022videoswin}
Ze Liu, Jia Ning, Yue Cao, Yixuan Wei, Zheng Zhang, Stephen Lin, and Han Hu.
\newblock Video swin transformer.
\newblock In \emph{CVPR}, 2022.

\bibitem[Loshchilov and Hutter(2019)]{loshchilov2019adamw}
Ilya Loshchilov and Frank Hutter.
\newblock Decoupled weight decay regularization.
\newblock In \emph{ICLR}, 2019.

\bibitem[Maes et~al.(2025)Maes, Le~Lidec, Scieur, LeCun, and
  Balestriero]{maes2025lejepa}
Luc Maes, Quentin Le~Lidec, Damien Scieur, Yann LeCun, and Randall Balestriero.
\newblock {LeJEPA}: Stable joint-embedding predictive architectures with
  sketched-isotropic-{G}aussian regularization.
\newblock \emph{arXiv preprint arXiv:2511.08544}, 2025.

\bibitem[Misra et~al.(2016)Misra, Zitnick, and Hebert]{misra2016shuffle}
Ishan Misra, C.~Lawrence Zitnick, and Martial Hebert.
\newblock Shuffle and learn: Unsupervised learning using temporal order
  verification.
\newblock In \emph{ECCV}, 2016.

\bibitem[Mur-Labadia et~al.(2026)Mur-Labadia, Muckley, Bar, Assran, Sinha,
  Rabbat, LeCun, Ballas, and Bardes]{murlabadia2026vjepa21}
Lorenzo Mur-Labadia, Matthew Muckley, Amir Bar, Mido Assran, Koustuv Sinha,
  Mike Rabbat, Yann LeCun, Nicolas Ballas, and Adrien Bardes.
\newblock {V-JEPA 2.1}: Unlocking dense features in video self-supervised
  learning.
\newblock \emph{arXiv preprint arXiv:2603.14482}, 2026.

\bibitem[Noroozi and Favaro(2016)]{noroozi2016jigsaw}
Mehdi Noroozi and Paolo Favaro.
\newblock Unsupervised learning of visual representations by solving jigsaw
  puzzles.
\newblock In \emph{ECCV}, 2016.

\bibitem[Oord et~al.(2018)Oord, Li, and Vinyals]{oord2018cpc}
Aaron van~den Oord, Yazhe Li, and Oriol Vinyals.
\newblock Representation learning with contrastive predictive coding, 2018.

\bibitem[Pan et~al.(2021)Pan, Song, Yang, Jiang, and Liu]{pan2021videomoco}
Tian Pan, Yibing Song, Tianyu Yang, Wenhao Jiang, and Wei Liu.
\newblock Videomoco: Contrastive video representation learning with temporally
  adversarial examples.
\newblock In \emph{CVPR}, 2021.

\bibitem[Pathak et~al.(2017)Pathak, Girshick, Doll{\'a}r, Darrell, and
  Hariharan]{pathak2017motion}
Deepak Pathak, Ross Girshick, Piotr Doll{\'a}r, Trevor Darrell, and Bharath
  Hariharan.
\newblock Learning features by watching objects move.
\newblock In \emph{CVPR}, 2017.

\bibitem[Recasens et~al.(2021)Recasens, Luc, Alayrac, Wang, Strub, Tallec,
  Malinowski, P{\u{a}}tr{\u{a}}ucean, Altch{\'e}, Valko, Grill, van~den Oord,
  and Zisserman]{recasens2021brave}
Adri{\`a} Recasens, Pauline Luc, Jean-Baptiste Alayrac, Luyu Wang, Florian
  Strub, Corentin Tallec, Mateusz Malinowski, Viorica P{\u{a}}tr{\u{a}}ucean,
  Florent Altch{\'e}, Michal Valko, Jean-Bastien Grill, Aaron van~den Oord, and
  Andrew Zisserman.
\newblock {Broaden Your Views for Self-Supervised Video Learning}.
\newblock In \emph{ICCV}, 2021.

\bibitem[Russakovsky et~al.(2015)Russakovsky, Deng, Su, Krause, Satheesh, Ma,
  Huang, Karpathy, Khosla, Bernstein, Berg, and
  Fei-Fei]{russakovsky2015imagenet}
Olga Russakovsky, Jia Deng, Hao Su, Jonathan Krause, Sanjeev Satheesh, Sean Ma,
  Zhiheng Huang, Andrej Karpathy, Aditya Khosla, Michael Bernstein,
  Alexander~C. Berg, and Li Fei-Fei.
\newblock {ImageNet} large scale visual recognition challenge.
\newblock In \emph{International Journal of Computer Vision}, 2015.

\bibitem[Soomro et~al.(2012)Soomro, Zamir, and Shah]{soomro2012ucf101}
Khurram Soomro, Amir~Roshan Zamir, and Mubarak Shah.
\newblock {UCF101}: A dataset of 101 human actions classes from videos in the
  wild.
\newblock In \emph{CRCV-TR-12-01}, 2012.

\bibitem[Sun et~al.(2019)Sun, Myers, Vondrick, Murphy, and
  Schmid]{sun2019contrastive}
Chen Sun, Austin Myers, Carl Vondrick, Kevin Murphy, and Cordelia Schmid.
\newblock Videobert: A joint model for video and language representation
  learning.
\newblock In \emph{ICCV}, 2019.

\bibitem[Tian et~al.(2020)Tian, Krishnan, and Isola]{tian2020cmc}
Yonglong Tian, Dilip Krishnan, and Phillip Isola.
\newblock Contrastive multiview coding.
\newblock In \emph{ECCV}, 2020.

\bibitem[Tong et~al.(2022)Tong, Song, Wang, and Wang]{tong2022videomae}
Zhan Tong, Yibing Song, Jue Wang, and Limin Wang.
\newblock {VideoMAE}: Masked autoencoders are data-efficient learners for
  self-supervised video pre-training.
\newblock In \emph{NeurIPS}, 2022.

\bibitem[Tran et~al.(2018)Tran, Wang, Torresani, Ray, LeCun, and
  Paluri]{tran2018r3d}
Du Tran, Heng Wang, Lorenzo Torresani, Jamie Ray, Yann LeCun, and Manohar
  Paluri.
\newblock A closer look at spatiotemporal convolutions for action recognition.
\newblock In \emph{CVPR}, 2018.

\bibitem[Vaswani et~al.(2017)Vaswani, Shazeer, Parmar, Uszkoreit, Jones, Gomez,
  Kaiser, and Polosukhin]{vaswani2017attention}
Ashish Vaswani, Noam Shazeer, Niki Parmar, Jakob Uszkoreit, Llion Jones,
  Aidan~N. Gomez, Lukasz Kaiser, and Illia Polosukhin.
\newblock Attention is all you need.
\newblock In \emph{NeurIPS}, 2017.

\bibitem[Wang et~al.(2023)Wang, Huang, Zhao, Tong, He, Wang, Wang, and
  Qiao]{wang2023videomaev2}
Limin Wang, Bingkun Huang, Zhiyu Zhao, Zhan Tong, Yinan He, Yi Wang, Yali Wang,
  and Yu Qiao.
\newblock {VideoMAE V2}: Scaling video masked autoencoders with dual masking.
\newblock In \emph{CVPR}, 2023.

\bibitem[Xie et~al.(2022)Xie, Zhang, Cao, Lin, Bao, Yao, Dai, and
  Hu]{xie2022simmim}
Zhenda Xie, Zheng Zhang, Yue Cao, Yutong Lin, Jianmin Bao, Zhuliang Yao, Qi
  Dai, and Han Hu.
\newblock {SimMIM}: A simple framework for masked image modeling.
\newblock In \emph{CVPR}, 2022.

\bibitem[Xu et~al.(2019)Xu, Xiao, Zhao, Shao, Xie, and Zhuang]{xu2019speednet}
Dejing Xu, Jun Xiao, Zhou Zhao, Jian Shao, Di Xie, and Yueting Zhuang.
\newblock Self-supervised spatiotemporal learning via video clip order
  prediction.
\newblock In \emph{CVPR}, 2019.

\bibitem[Yan et~al.(2022)Yan, Xiong, Arnab, Lu, Zhang, Sun, and
  Schmid]{yan2022multiview}
Shen Yan, Xuehan Xiong, Anurag Arnab, Zhicheng Lu, Mi Zhang, Chen Sun, and
  Cordelia Schmid.
\newblock Video representation learning using discriminative pooling.
\newblock In \emph{CVPR}, 2022.

\bibitem[Yu et~al.(2025)Yu, Chen, Liu, Horiuchi, Braverman, Xu, Haramati, and
  Balestriero]{yu2025auxjepa}
Jiacan Yu, Siyi Chen, Mingrui Liu, Nono Horiuchi, Vladimir Braverman, Zicheng
  Xu, Dan Haramati, and Randall Balestriero.
\newblock Why and how auxiliary tasks improve {JEPA} representations.
\newblock \emph{arXiv preprint arXiv:2509.12249}, 2025.

\bibitem[Zbontar et~al.(2021)Zbontar, Jing, Misra, LeCun, and
  Deny]{zbontar2021barlow}
Jure Zbontar, Li Jing, Ishan Misra, Yann LeCun, and St{\'e}phane Deny.
\newblock Barlow twins: Self-supervised learning via redundancy reduction.
\newblock In \emph{ICML}, 2021.

\end{thebibliography}
}

\clearpage

\appendix

\section{Pseudocode for Key Methods}
\label{sec:pseudocode}

Algorithm~\ref{alg:baseline} summarizes the standard V-JEPA training loop. Algorithm~\ref{alg:fwmhwld} describes FWM-HW-LD, which extends the baseline with hard-weighted prediction, factorization losses, and hard-weighted latent dynamics.

\begin{algorithm}[H]
\caption{Baseline V-JEPA Training}
\label{alg:baseline}
\begin{algorithmic}[1]
\REQUIRE Video clip $x$, random mask $m$
\STATE $z_{\text{vis}} \gets f_\theta(x, \text{visible}=\overline{m})$
\STATE $\hat{z} \gets g_\phi(z_{\text{vis}}, m)$
\STATE $h \gets \text{sg}(f_{\bar{\theta}}(x))$
\STATE $\mathcal{L} \gets \|\hat{z} - h[m]\|_1$
\STATE Update $f_\theta, g_\phi$ via backprop
\STATE EMA: $\bar{\theta} \gets 0.99925 \cdot \bar{\theta} + 0.00075 \cdot \theta$
\end{algorithmic}
\end{algorithm}

\begin{algorithm}[H]
\caption{FWM-HW-LD Training (Eq.~\ref{eq:fwmhwld})}
\label{alg:fwmhwld}
\begin{algorithmic}[1]
\REQUIRE Video clip $x$, mask $m$, ratios $D_{\text{app}}{=}D/2$
\STATE Compute JEPA loss $\mathcal{L}_{\text{JEPA}}$ (Algorithm~\ref{alg:baseline})
\STATE $e^{\text{jepa}}_i \gets \|\hat{z}_i - h_i[m]\|_1$
\STATE $w^{\text{jepa}}_i \gets \text{softmax}(e^{\text{jepa}}_i / \tau) \cdot N_{\text{tok}}$
\STATE $\mathcal{L}_{\text{HW-JEPA}} \gets \text{mean}(w^{\text{jepa}}_i \cdot e^{\text{jepa}}_i)$
\STATE $z \gets f_\theta(x)$, \quad $h \gets \text{sg}(f_{\bar{\theta}}(x))$
\STATE $Z_{\text{app}}, Z_{\text{dyn}} \gets z[\ldots, {:}D_{\text{app}}], \; z[\ldots, D_{\text{app}}{:}]$
\STATE $\mathcal{L}_{\text{static}} \gets \text{mean}(|Z_{\text{app}}^{(t+1)} - Z_{\text{app}}^{(t)}|)$
\STATE $\mathcal{L}_{\text{orth}} \gets \|C_{Z_{\text{app}}}^\top C_{Z_{\text{dyn}}}\|_F^2 / N$
\STATE $\hat{\Delta} \gets \text{DynHead}(Z_{\text{dyn}}^{(t)})$
\STATE $\Delta h \gets h^{(t+1)} - h^{(t)}$
\STATE $e_i \gets \|\hat{\Delta}_i - \Delta h_i\|_1$ \COMMENT{per-token error}
\STATE $w^{\text{ld}}_i \gets \text{softmax}(e_i / \tau) \cdot N_{\text{tok}}$ \COMMENT{hard-region weights}
\STATE $\mathcal{L}_{\text{LD-HW}} \gets \text{mean}(w^{\text{ld}}_i \cdot e_i)$
\STATE $\mathcal{L} \gets \mathcal{L}_{\text{JEPA}} + \lambda_{hw}\mathcal{L}_{\text{HW-JEPA}} + \lambda_s \mathcal{L}_{\text{static}} + \lambda_o \mathcal{L}_{\text{orth}} + \lambda_d \mathcal{L}_{\text{LD-HW}}$
\end{algorithmic}
\end{algorithm}

The dynamics head \texttt{DynHead} is a two-layer MLP with hidden dimension 256 and GELU activation. When FWM is active, the input dimension is reduced to $D_{\text{dyn}} = D - D_{\text{app}}$ to ensure that dynamics prediction does not draw on appearance features.

\section{Hyperparameters}
\label{sec:hyperparams}

Table~\ref{tab:hyperparams} lists all hyperparameters across the 18 method variants. We did not perform per-method hyperparameter tuning; values were chosen based on small-scale preliminary runs and held fixed across all main experiments to ensure fair comparison. Coefficients $\lambda_s$ and $\lambda_o$ for FWM are deliberately small to act as regularizers rather than dominant objectives, while $\lambda_d$ is unit weight to provide a meaningful learning signal for the dynamics head. The HW temperature $\tau$ controls the sharpness of hard-region weighting; $\tau=1$ provides a soft attention over the per-token error distribution.

\begin{table}[H]
    \caption{Hyperparameters for all auxiliary objectives.}
    \label{tab:hyperparams}
    \centering
    \footnotesize
    \setlength{\tabcolsep}{4pt}
    \begin{tabular}{llc}
        \toprule
        \textbf{Parameter} & \textbf{Used by} & \textbf{Value} \\
        \midrule
        $\lambda_{\text{kin}}$ & Kinematic variants & 0.1 or 1.0 \\
        $\lambda_s$ (static) & FWM variants & 0.05 \\
        $\lambda_o$ (orth.) & FWM variants & 0.01 \\
        $D_{\text{app}} / D$ & FWM variants & 0.5 \\
        $\lambda_d$ (LD) & LD variants & 1.0 \\
        LD hidden dim & LD variants & 256 \\
        HW temperature $\tau$ & HW variants & 1.0 \\
        $\lambda_{\text{HW}}$ & HW, AC+HW, Combo / HW-LD, FWM-HW-LD & 0.3 / 1.0 \\
        $\lambda_{\text{AC}}$ & AC, AC+HW, FAC & 1.0 \\
        $\lambda_\Delta$ (delta) & Delta-JEPA, Combo & 0.5 \\
        $\lambda_{\text{spec}}$ & Spectral-JEPA & 1.0 \\
        $\lambda_{\text{LTC}}$ / margin & LTC-JEPA & 0.5 / 0.5 \\
        Motion fallback rate & Motion-Guided / AMG & 0.1 / 0.0 \\
        \bottomrule
    \end{tabular}
\end{table}

\section{Computational Resources}
\label{sec:compute}

All experiments were conducted on a Kubernetes cluster with NVIDIA A100 GPUs (40~GB). Each pretraining run uses 4~GPUs with global batch size 32 for 100 epochs of 300 iterations each, completing in approximately 7 hours. Frozen-encoder evaluation runs require an additional 2--6 hours per benchmark, depending on dataset size. Table~\ref{tab:compute} summarizes the total budget.

\begin{table}[H]
    \caption{Computational budget for the full study.}
    \label{tab:compute}
    \centering
    \footnotesize
    \begin{tabular}{lc}
        \toprule
        \textbf{Resource} & \textbf{Value} \\
        \midrule
        GPUs per training run & $4\times$ NVIDIA A100 (40\,GB) \\
        Training time per run & $\sim$7 hours \\
        Number of training runs & 18 \\
        Total pretraining GPU-h & $\sim$504\,GPU-h \\
        Eval runs per method & 3 (D-48, IN-100, SSv2) \\
        Eval time per benchmark & $\sim$2--6 hours \\
        Total evaluation GPU-h & $\sim$720\,GPU-h \\
        \midrule
        Total compute & $\sim$1{,}224\,GPU-h \\
        Backbone & ViT-Base (86M parameters) \\
        Framework & PyTorch, distributed (DDP) \\
        \bottomrule
    \end{tabular}
\end{table}

\section{Additional Implementation Notes}
\label{sec:implementation}

\noindent\textbf{Mixed-data sampling.} For mixed-dataset pretraining, we sample batches with weights 20\% UCF-101, 60\% SSv2, and 20\% ImageNet-100. Image samples from ImageNet-100 are converted to single-frame ``videos'' (tubelet length 1) so that the same data pipeline handles all three sources. The 60\% SSv2 weight reflects our explicit goal of strengthening temporal reasoning.

\noindent\textbf{Mask sampling.} The default V-JEPA mask is a multi-block spatiotemporal tube. For Motion-Guided Masking, we compute per-frame motion energy as the mean L1 difference between consecutive frames at the patch level, normalize per-clip, and sample mask centers proportional to the resulting score. With probability 0.1 we fall back to uniform random sampling to retain coverage of static regions.

\noindent\textbf{Latent dynamics head input.} Under FWM-HW-LD, the dynamics head receives only the $D_{\text{dyn}} = D - D_{\text{app}}$ dynamic channels of the student representation; this is implemented by a tensor slice prior to the linear projection. Removing this slice (i.e., letting the dynamics head see the full $D$-dimensional representation) recovers the LD-JEPA baseline and substantially weakens appearance preservation.

\noindent\textbf{Numerical stability.} The hard-region weighting softmax operates on per-token L1 errors that can vary by orders of magnitude across batches. We clip the temperature-scaled logits to $[-20, 20]$ before the softmax to prevent numerical overflow, and we detach the weights from the autograd graph (\texttt{torch.no\_grad}) so that gradients flow only through the unweighted error.

\end{document}